\documentclass[runningheads]{llncs}
\usepackage[T1]{fontenc}
\usepackage{graphicx,verbatim}
\usepackage{multirow}
\usepackage{bbding}
\usepackage{pifont}
\usepackage{amssymb}
\usepackage{booktabs}
\usepackage[table,xcdraw]{xcolor}
\begin{document}
\title{Boosting Generalizable Depth Estimation in Endoscopy by Mixture of Lightweight Experts and Intrinsic Image Alignment}
%
\author{Liangjing Shao\inst{1,2} \and
Beilei Cui\inst{1} \and Yiming Huang\inst{1} \and Changjing Liu\inst{1} \and \\
Hongliang Ren\inst{\footnotemark[1],1,2}}
\authorrunning{L. Shao et al.}
%
\institute{Department of Electronic Engineering, The Chinese University of Hong Kong, Hong Kong SAR, China \and
Shenzhen Loop Area Institute, China \\
\email{leonking-shaw@link.cuhk.edu.hk, hlren@ee.cuhk.edu.hk}}

\footnotetext[1]{Corresponding Author}


\maketitle              
\begin{abstract}
Depth estimation is a significant task for 3D perception in endoscopic surgeries. However, illumination interference and feature diversity in various endoscopic scenes are still challenges for generalizable depth estimation and ego-motion estimation. Based on this, a novel self-supervised framework, EndoMINI, is proposed for depth estimation in endoscopic scenes. Specifically, mixture of low-rank experts (MiLoRE) is proposed to perform parameter-efficient fine-tuning, which can also boost the model adaptation to scenes with different characteristics. Meanwhile, an intrinsic image alignment (IIA) is introduced into the training loss to alleviate the influence of light reflectance in endoscopy with a novel intrinsic image decomposition network. The proposed method is evaluated on SCARED datasets for supervised depth estimation, and two endoscopic datasets, Hamlyn and SERV-CT, for zero-shot depth estimation, compared with state-of-the-art works as well. The experimental results demonstrate outstanding performance of the proposed model and the effects of the main contributions.


\end{abstract}
\section{Introduction}
\begin{figure}
    \centering
    \includegraphics[width=0.9\linewidth]{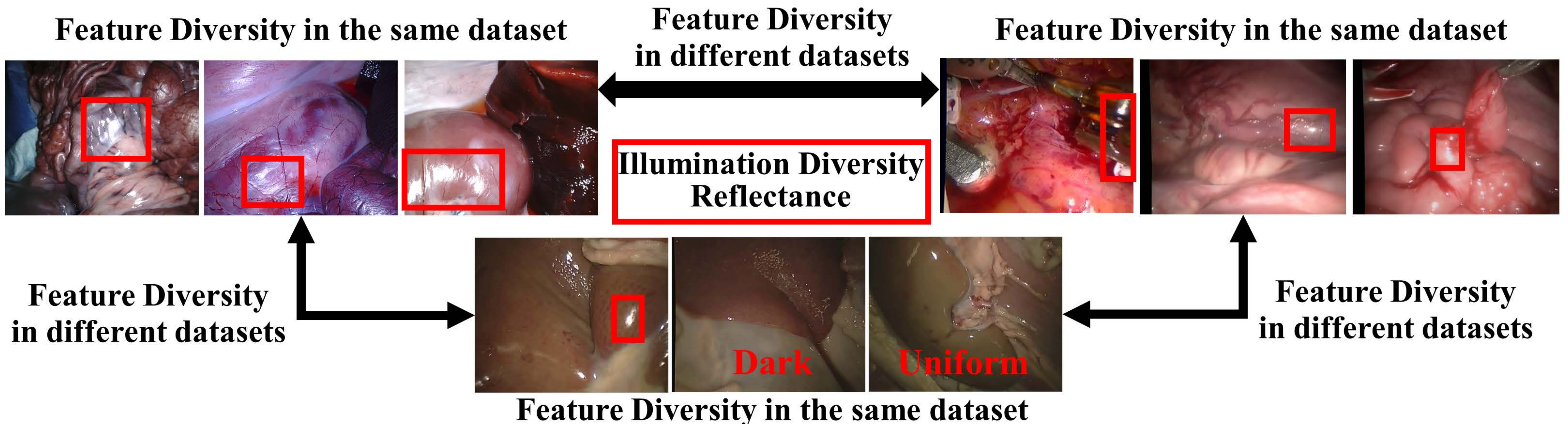}
    \caption{The diversity in endoscopic scenes.}
    \label{mot}
\end{figure}

Depth estimation has been a vital task for minimally invasive endoscopic surgery, especially for robotic surgery, which plays an important role in navigation and 3D perception.\cite{bg} A series of self-supervised learning frameworks have become the mainstream methods for depth estimation in natural scenes\cite{md2,mv,lm}. However, features of endoscopic scenes are quite different from natural scenes. Meanwhile, illumination inconsistency is also another critical problem in endoscopic scenes. Currently, a large amount of novel frameworks are proposed for depth estimation in endoscopic scenes, which tackle these problems with various alignment losses\cite{afsfm,iidsfm,dvs}, parameter-efficient fine-tuning(PEFT)\cite{endodac}, and efficient feature encoders\cite{tmi,endoslam}. However, there are still two challenges for existing methods as Fig. \ref{mot} shows. Firstly, for endoscopic data, even from the same dataset, the features of endoscopic scenes are quite different due to various characteristics of tissues. Therefore, feature encoders usually suffer from corresponding representation extraction for various scenes and objects. Furthermore, although the existing method\cite{afsfm} alleviates the illumination inconsistency with appearance flow and optical flow, the reflectance on the surface of tissues in different illumination conditions still affects the accuracy of depth estimation. 

Recently, some remarkable works are proposed to deal with these two challenges. In EndoDAC\cite{endodac}, a novel LoRA module is proposed to perform PEFT on Depth Anything\cite{da}, a general depth estimation model. However, single LoRA layer still faces the challenge of feature adaptation for various scenes. Moreover, a recent work, IID-SfMLearner\cite{iidsfm}, proposed to utilize intrinsic image decomposition to deal with the reflectance problem in endoscopy. However, the style gap between decomposed images and illumination inconsistency are ignored in their proposed framework. Currently, mixture of experts (MoE)\cite{moe-r} has been an excellent parameter-efficient fine-tuning method, which can adjust the activation of different experts according to the features from various data. Therefore, to tackle the problem from various endoscopic characteristics, a novel PEFT module is proposed, in which LoRA layers are used as experts in a MoE framework. Furthermore, in this work, an intrinsic image alignment based on a novel self-supervised style-controlled intrinsic image decomposition is proposed and integrated with reconstruction alignments based on optical flow and appearance flow\cite{afsfm}. To this end, our main contributions are summarized as the following:
\begin{enumerate}
    \item A novel parameter-efficient fine-tuning framework with integration of MoE and LoRA (MiLoRE) is proposed to boost the generalization of the model for endoscopic scenes with various characteristics.
    \item With a novel self-supervised intrinsic image decomposition network, intrinsic image alignment including an illumination-free alignment and an intrinsic image reconstruction alignment is introduced into the training stage to alleviate the influence of various light conditions.
    \item The depth estimation network based on self-supervised learning with the proposed framework outperforms state-of-the-art methods on SCARED dataset for supervised depth estimation, Hamlyn and SERV-CT dataset for zero-shot depth estimation. Besides, the proposed framework provides the most accurate ego-motion estimation and camera intrinsics prediction.
\end{enumerate}

\section{Proposed Method}
\subsection{Self-supervised Pipeline}
\begin{figure}
    \centering
    \includegraphics[width=\linewidth]{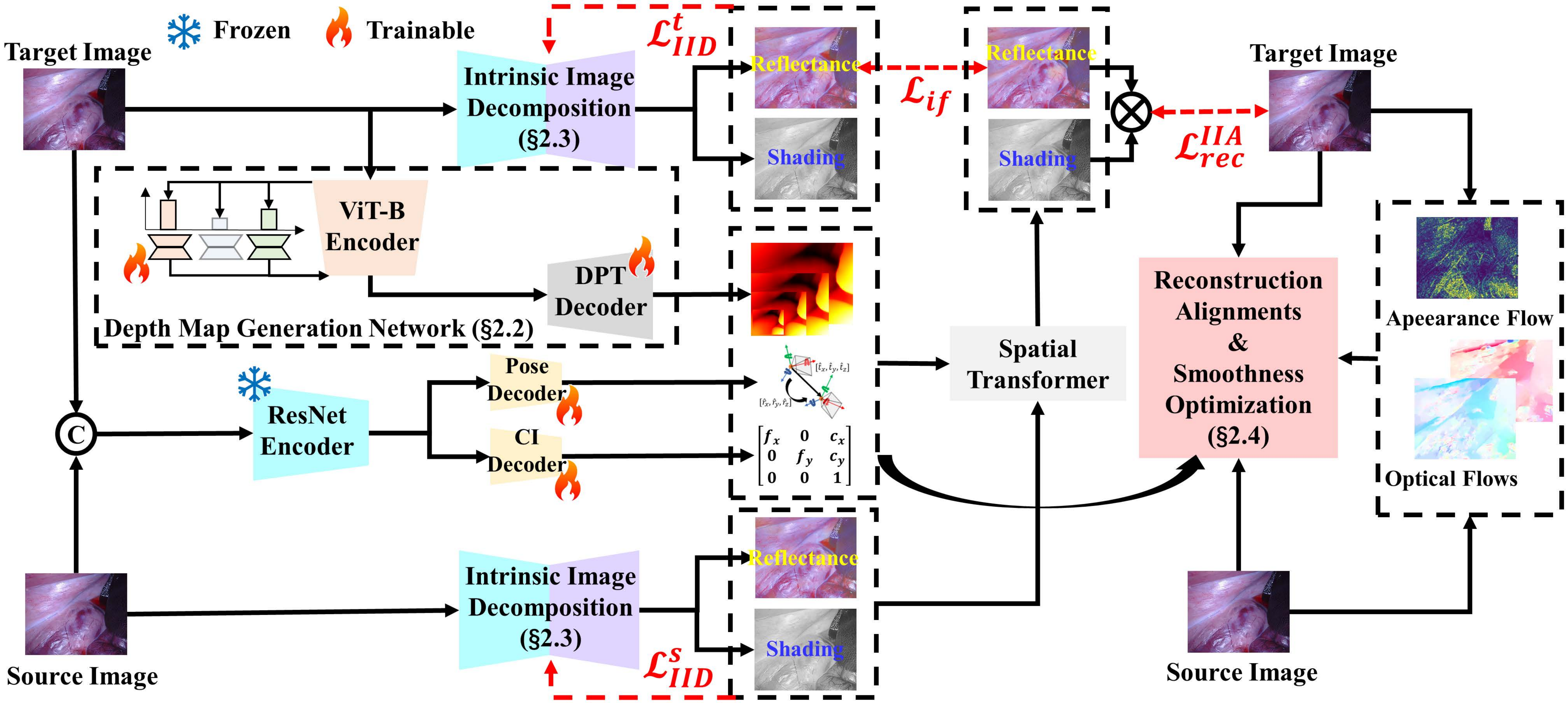}
    \caption{Self-supervised pipeline for endoscopy depth estimation.}
    \label{ppl}
\end{figure}
The proposed pipeline in this work is shown in Fig. \ref{ppl}. The depth map generation network in the proposed framework mainly consists of a pretrained ViT encoder with convolution blocks inserted and a DPT-like decoder from EndoDAC\cite{endodac} and Depth Anything\cite{da}. Specifically, a novel mixture of low-rank experts is proposed to perform parameter-efficient fine-tuning on each Transformer block in the pretrained ViT encoder. Apart from existing reconstruction alignments and smoothness optimization\cite{afsfm}, an illumination-free alignment and an intrinsic image reconstruction alignment based on intrinsic image decomposition are introduced to optimize the depth estimation and ego-motion estimation. Besides, a novel self-supervised intrinsic image decomposition network is proposed to obtain accurate reflectance images and shading images for the illumination-free alignment and the intrinsic image reconstruction alignment.

\subsection{Mixture of Low-Rank Experts}
\begin{figure}
    \centering
    \includegraphics[width=\linewidth]{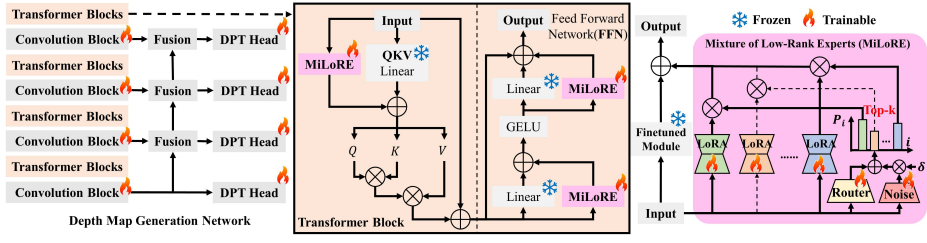}
    \caption{Parameter-efficient finetuning on each Transformer block in ViT-B encoder based on mixture of low-rank experts.}
    \label{milore}
\end{figure}
In the proposed mixture of low-rank experts, the low-rank adaptation modules can adaptively adjust the output of the finetuned module based on the input data. To achieve this, the probability distribution of experts is generated by a router network $\mathcal{R}$ based on the input feature at first. Based on this, $k$ experts with highest probability are selected to perform weighted Low-Rank adaptation (LoRA\cite{lora}). For the finetuned module with the frozen weight $W_0\in\mathbf{R}^{d\times k}$, given the input $x$, the output $y$ is adjusted by selected $k$ low-rank experts as Eq. \ref{peft} shows. In Eq.\ref{peft}, $B_i\in\mathbb{R}^{d\times r}$ and $A_i\in\mathbb{R}^{r\times k}$ are trainable weights, $r\ll min(d,k)$. To increase the robustness of the router, a noise vector $\epsilon(x)$ is predicted based on the input and integrated with a random noise vector $\delta$.
\begin{equation}
    y =Wx= W_0x+\sum_{i=1}^{k}(\mathcal{R}(x)+\epsilon(x)\times\delta)_iB_iA_ix
\label{peft}
\end{equation}

To finetune the pretrained ViT encoder for adaptation to the endoscopic scenes, three modules based on MiLoRE are implemented in each Transformer block. As Fig. \ref{milore} shows, one MiLoRE module is utilized to finetune 'QKV Linear' (QKV) which generates query vector, key vector and value vector for self-attention operation. Moreover, another two MiLoRE modules are set to respectively finetune two MLPs in the feed forward network (FFN) of the block.

\subsection{Intrinsic Image Decomposition}
\begin{figure}
    \centering
    \includegraphics[width=\linewidth]{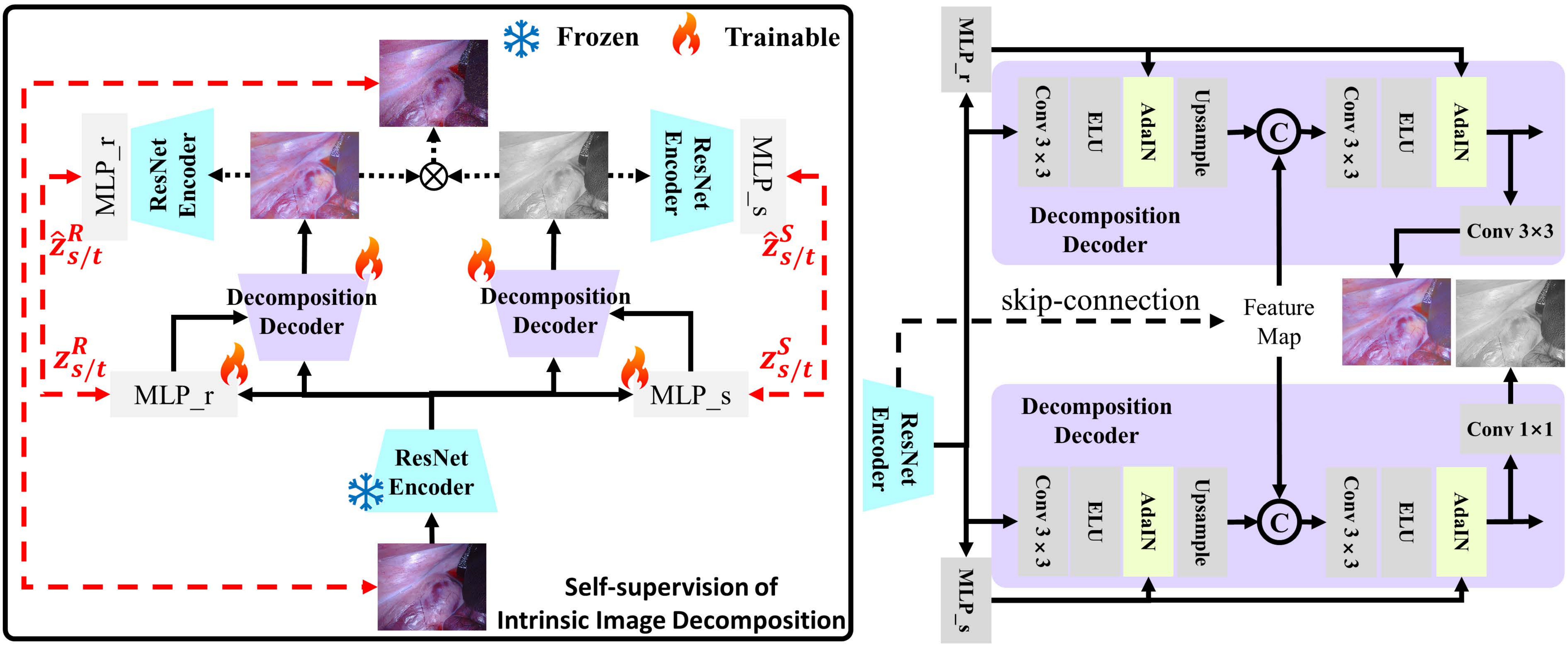}
    \caption{The framework of the proposed intrinsic image decomposition network.}
    \label{iia}
\end{figure}
The pipeline of the proposed intrinsic image decomposition network is shown in Fig.\ref{iia}. Due to the style gap between the reflectance image and the shading image, two MLPs are introduced into the proposed intrinsic image decomposition network to generate the style vector $z^R$ and $z^S$, respectively. To control the style of the generated image, referring to \cite{iid},  Adaptive Instance Normalization (AdaIN) is utilized instead of original normalization layer in the decomposition decoder. The parameters of AdaIN are from the corresponding style vector dynamically generated by the specific MLP. The process of AdaIN is shown as Eq. \ref{adain}, in which $a$ is the output of the activation layer, $z=[\beta,\gamma]$ is the dynamic parameter of AdaIN, $\mu$ and $\sigma$ denote channel wise mean and standard deviation.
\begin{equation}
    AdaIN(a,\beta,\gamma)=\gamma\frac{a-\mu(a)}{\sigma(a)}+\beta
\label{adain}
\end{equation}

In the pipeline, two intrinsic image decomposition decoders generate the corresponding reflectance images $R_t, R_s$ and shading images $S_t,S_s$ from the source image and target image, respectively. The style vectors $\hat{z}^R$ and $\hat{z}^S$ are also predicted from the decomposed images. The masked photometric loss between the input image and the composed image with the L1 loss of corresponding style vectors is calculated to optimize the decoders directly, shown as Eq. \ref{L_iid}.
\begin{equation}
    \mathcal{L}_{IID}^{s/t} = \mathcal{L}_p(I_{s/t},R_{s/t}\otimes S_{s/t}) + |z^R_{s/t}-\hat{z}^R_{s/t}|_1 + |z^S_{s/t}-\hat{z}^S_{s/t}|_1
\label{L_iid}
\end{equation}
where the masked photometric loss is defined as the Eq. \ref{lp}, in which $M_v$ is the visibility mask generated based on backward optical flow \cite{afsfm}, $\alpha=0.85$.
\begin{equation}
    \mathcal{L}_p(I_1,I_2)=M_v(\alpha\frac{1-SSIM(I_1,I_2)}{2}+(1-\alpha)|I_1-I_2|)
    \label{lp}
\end{equation}

\subsection{Loss Functions for Depth Estimation}
Based on the rigid transformation from the predicted depth map $D_t$ and the ego-motion, the reflectance image and the shading image from the source image can be warped into the corresponding decomposed images $R_{s\rightarrow t}$ and $S_{s\rightarrow t}$, which are expected to be the same as the decomposed image from the target image. Therefore, the masked photometric loss between $R_{s\rightarrow t}$ and $R_t$ in Eq. \ref{if} is utilized to perform illumination-free alignment, as the illumination condition will not influence the reflectance image. Based on optical flow estimation and appearance flow $C_{s,t}$ estimation in \cite{afsfm}, the alignments between the target image without light inconsistency $I'_t=I_t+C_{s,t}$ and the reconstructed images from the optical flow $I^{op}_{s\rightarrow t}$, the depth map $I^{de}_{s\rightarrow t}$ and the intrinsic images $R_{s\rightarrow t}\otimes S_{s\rightarrow t}$ can be performed using the loss function Eq. \ref{rec}. Moreover, the smoothness loss for appearance flow and depth map can be obtained and integrated for the smoothness alignment by Eq. \ref{sa}.
\begin{equation}
    \mathcal{L}_{if} = \mathcal{L}_p(R_{s\rightarrow t}, R_t)
\label{if}
\end{equation}
\begin{equation}
    \mathcal{L}_{rec} = \lambda_4\mathcal{L}_p(I^{op}_{s\rightarrow t}, I'_t)+\lambda_5\mathcal{L}_p(I^{de}_{s\rightarrow t}, I'_t)+\lambda_6\mathcal{L}_p(R_{s\rightarrow t}\otimes S_{s\rightarrow t}, I'_t)
\label{rec}
\end{equation}
\begin{equation}
    \mathcal{L}_{sa} = \nabla |C_{s,t}| \cdot e^{\nabla|I_t - I^{op}_{s\rightarrow t}|} + \nabla |D_t| \cdot e^{\nabla|I_t|}
\label{sa}
\end{equation}
The whole loss function $\mathcal{L}$ is calculated by Eq. \ref{loss}, in which $\lambda_1=0.01$, $\lambda_2=0.02$, $\lambda_3=0.001$, $\lambda_4=0.01$, $\lambda_5=0.01$, $\lambda_6=0.1$ according to the scale of each loss.
\begin{equation}
    \mathcal{L}=\lambda_1(\mathcal{L}_{IID}^{s}+\mathcal{L}_{IID}^{t})+\lambda_2\mathcal{L}_{if}+\lambda_3\mathcal{L}_{sa}+\mathcal{L}_{rec}
    \label{loss}
\end{equation}

\section{Experiments and Results}

\subsection{Implementations and Datasets}
The proposed framework is implemented on one NVIDIA RTX 4090 GPU. The rank of LoRA experts is set to 4. The number of LoRA experts for the FFN is set to 5, which for QKV is set to 3. Top-k is set to 2 for both FFN and QKV. The learning rate of Adam optimizer is set to 1e-4 initially, which is scaled by a factor of 0.1 after 10 epochs. The framework is trained for 20 epochs in total with a batch size of 8. Following most of the existing works \cite{endodac},\cite{afsfm}, five metrics including Absolute Relateive Error($Rel_{Abs}$), Squared Absolute Relative Error($Rel_{Sq}$), Root Mean Square Error($RMSE$) and its logarithmic term($RMSE_{Log}$) are used for evaluation. Three public endoscopic datasets are utilized for the experiments, including SCARED\cite{scared}, Hamlyn\cite{hamlyn} and SERV-CT\cite{servct}, which are collected by da Vinci robotic systems. The resolution of all images are resized to $320\times256$.

\textbf{SCARED dataset}: Following \cite{endodac} and \cite{afsfm}, the dataset is split into 15351, 1705, and 551 frames for training, validation and evaluation, respectively.

\textbf{Hamlyn dataset}: Following \cite{endodac}, all 92672 frames from 21 rectified videos \cite{rec} are selected for for evaluation of zero-shot depth estimation.

\textbf{SERV-CT dataset}: Following \cite{dvs}, all of 32 keyframes are utilized for zero-shot depth estimation in the experiments.

\begin{table}[!ht]
\centering
\caption{Results on SCARED dataset and \textbf{Zero-shot} results on Hamlyn, SERV-CT. \textbf{TP}: the number of \textbf{trainable parameters} in the depth map generation network.}
{\fontsize{7}{8}\selectfont
\begin{tabular}{c|c|c|ccccc|c}
\hline
\textbf{}                                            & \textbf{Methods}                        & \textbf{Year} & \textbf{$Rel_{Abs}\downarrow$} & \textbf{$Rel_{Sq}\downarrow$} & \textbf{$RMSE\downarrow$} & \textbf{$RMSE_{Log}\downarrow$} & \textbf{$\delta\uparrow$} & \textbf{TP}(M)\\ \hline
\multirow{6}{*}{{\rotatebox[origin=c]{90}{SCARED}}}  & Depth Anything$\ddagger$ \cite{da}      & 2024       & 0.055                          & 0.410                         & 4.769                     & 0.078                           & 0.973           &     13.0     \\
                                                     & Depth Anything v2$\ddagger$ \cite{da2}    & 2024    & 0.076                          & 0.683                         & 6.379                     & 0.104                           & 0.949      &    13.0           \\
                                                     & IID-SfMLearner \cite{iidsfm}             & 2024       & 0.058                          & 0.435                         & 4.820                     & 0.080                           & 0.969               &   14.8   \\
                                                     & DVSMono$\blacktriangle$ \cite{dvs}                               & 2024      & 0.055                          & 0.410                         & 4.797                     & 0.078                           & 0.975      &    27.0           \\
                                                     & EndoDAC\ding{72} \cite{endodac}          & 2024     & \underline{0.052}                          & \underline{0.362}                         & \underline{4.464}                     & \underline{0.073}                           & \underline{0.979}              &     \textbf{1.6}  \\
                                                    & MonoPCC\cite{pcc}                                              & 2025                     & 0.051                          & 0.349                         & 4.488                     & 0.072                           & 0.983  & 27.0 \\                                
                                                     & {\cellcolor[HTML]{EFEFEF}EndoMINI(Ours)\ding{72} }               & {\cellcolor[HTML]{EFEFEF}2025}          &    {\cellcolor[HTML]{EFEFEF}\textbf{0.047}}            &     {\cellcolor[HTML]{EFEFEF}\textbf{0.315}}           &      {\cellcolor[HTML]{EFEFEF}\textbf{4.246}}       &          {\cellcolor[HTML]{EFEFEF}\textbf{0.067}}         &   {\cellcolor[HTML]{EFEFEF}\textbf{0.984}}   &   {\cellcolor[HTML]{EFEFEF}\underline{3.6}}   \\ \hline
\multirow{6}{*}{{\rotatebox[origin=c]{90}{Hamlyn}}}  & Depth Anything$\ddagger$ \cite{da}      & 2024       & 0.154                          & 3.616                         & 12.733                    & 0.189                           & 0.784           &     13.0     \\
                                                     & Depth Anything v2$\ddagger$ \cite{da2}             & 2024       & 0.182                          & 4.994                         & 15.067                    & 0.219                           & 0.740       &       13.0       \\
                                                     & IID-SfMLearner$\dagger$ \cite{iidsfm}    & 2024       & 0.171                          & 4.526                         & 14.066                    & 0.206                           & 0.767        &       14.8      \\
                                                     & DVSMono$\blacktriangle$ \cite{dvs}              & 2024      & \underline{0.143}                          & \textbf{2.956}                         & \underline{11.905}                   & \underline{0.181}                           & \underline{0.796}        &      27.0       \\
                                                     & EndoDAC\ding{72} \cite{afsfm}            & 2024     & 0.156                          & 3.854                         & 12.936               & 0.193                           & 0.791        &       \textbf{1.6}      \\
                                                     & MonoPCC\cite{pcc}                                              & 2025                     & 0.158                          & 3.889                         & 13.205                    & 0.194                           & 0.782  & 27.0 \\
                                                     & {\cellcolor[HTML]{EFEFEF}EndoMINI(Ours)\ding{72}}                  & {\cellcolor[HTML]{EFEFEF}2025}          & {\cellcolor[HTML]{EFEFEF}\textbf{0.140}}                 & {\cellcolor[HTML]{EFEFEF}\underline{3.087}}                & {\cellcolor[HTML]{EFEFEF}\textbf{11.700}}           & {\cellcolor[HTML]{EFEFEF}\textbf{0.174}}                  & {\cellcolor[HTML]{EFEFEF}\textbf{0.813}}       &   {\cellcolor[HTML]{EFEFEF}\underline{3.6}}   \\ \hline
\multirow{6}{*}{{\rotatebox[origin=c]{90}{SERV-CT}}} & Depth Anything$\ddagger$ \cite{da}                 & 2024   & \textbf{0.082}          & \underline{1.122}          & \underline{9.409}         & \underline{0.110}          & \underline{0.929}      &  13.0  \\
                                                     & Depth Anything v2$\ddagger$ \cite{da2}             & 2024   & 0.092          & 1.328          & 10.407         & 0.120          & 0.918    &    13.0  \\
                                                     & IID-SfMLearner$\dagger$ \cite{iidsfm} & 2024   & 0.112          & 1.956          & 12.193         & 0.138          & 0.878      &   14.8 \\
                                                     & DVSMono$\blacktriangle$ \cite{dvs}              & 2024  & 0.095          & 1.412          & 10.811         & 
                                                     0.127          & 0.907     &  27.0   \\
                                                     & EndoDAC$\dagger$\ding{72} \cite{endodac} & 2024 & 0.084 & 1.223          & 9.683          & 0.113          & 0.922      &  \textbf{1.6}  \\
                                                     & MonoPCC\cite{pcc}                                              & 2025                      & 0.091                          & 1.265                         & 10.123                    & 0.117                           & 0.915  & 27.0 \\
                                                     & {\cellcolor[HTML]{EFEFEF}EndoMINI(Ours)\ding{72}}                  & {\cellcolor[HTML]{EFEFEF}2025}      & {\cellcolor[HTML]{EFEFEF}\underline{0.083}} & {\cellcolor[HTML]{EFEFEF}\textbf{1.081}} & {\cellcolor[HTML]{EFEFEF}\textbf{8.948}} & {\cellcolor[HTML]{EFEFEF}\textbf{0.105}} & {\cellcolor[HTML]{EFEFEF}\textbf{0.935}} & {\cellcolor[HTML]{EFEFEF}\underline{3.6}}
                                                     \\ \hline
\multicolumn{9}{l}{\textbf{The best results} are in bold, \underline{the second best results} are underlined.} \\
\multicolumn{9}{l}{$\ddagger$: Finetuned on SCARED dataset with convolution blocks and self-supervision.} \\  
\multicolumn{9}{l}{$\dagger$: Without public results, the public weight from training on SCARED is used.} \\ 
\multicolumn{9}{l}{$\blacktriangle$: Due to different training splits, the public weight from the same split is used.} \\        
\multicolumn{9}{l}{\ding{72}: \textbf{Without} given camera intrinsics.} \\  \hline           
\end{tabular}}
\label{comp}
\end{table}

\subsection{Comparison Results}
\begin{figure}[!ht]
    \centering
    \includegraphics[width=\linewidth]{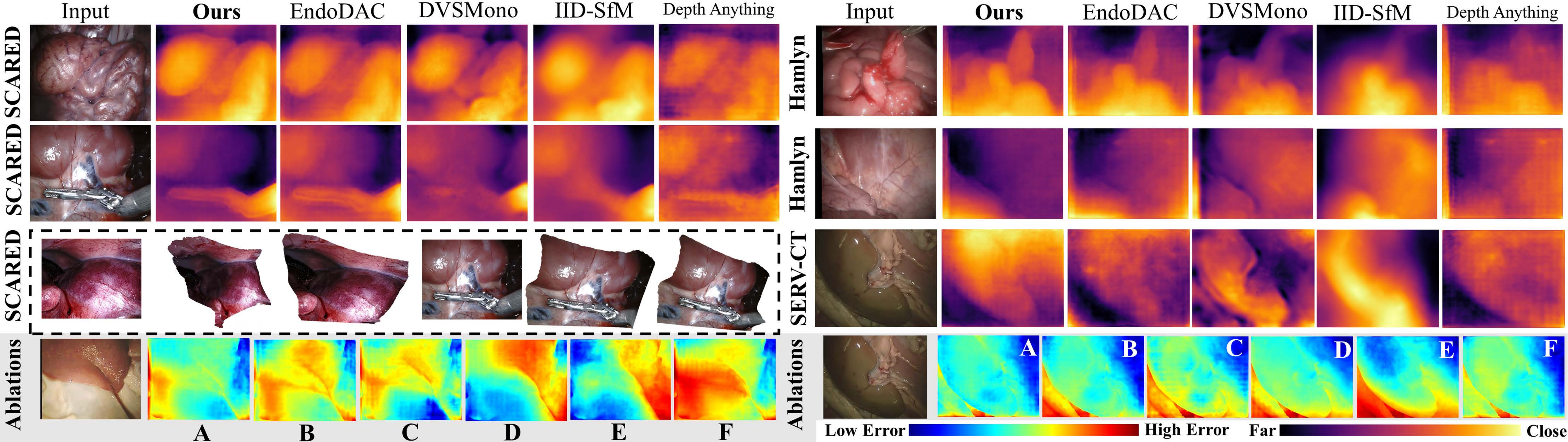}
    \caption{Visualization results. The corresponding ablation settings are in Table. \ref{ab}.}
    \label{vis}
\end{figure}

The proposed framework is compared with state-of-the-art works, which are with public codes or weights, on SCARED dataset for supervised depth estimation, while the evaluation of zero-shot depth estimation with models trained on SCARED dataset is also performed on Hamlyn and SERV-CT. As Table \ref{comp} shows, the proposed framework outperforms all of state-of-the-art methods on SCARED dataset for supervised depth estimation. Moreover, as Table \ref{comp} shows, the proposed framework can provide the most accurate zero-shot depth estimation on both of Hamlyn and SERV-CT dataset, which demonstrates the outstanding generalization of the proposed model. Besides, qualitative results from different datasets in Fig. \ref{vis} also demonstrate the excellent performance of our method, which also includes two 3D reconstruction samples based on our method in the dashed box. 


\subsection{Ablation Studies}
To demonstrate the effects of our main contributions, a series of ablation studies are performed in the experiments additionally. All of ablation results can be found in Table \ref{ab} and the last row in Fig. \ref{vis} (A is ours), which includes quantitative results on SCARED dataset and Hamlyn dataset, as well as qualitative results on SERV-CT dataset. Firstly, with evaluation based on varying numbers of experts in studies A-C, the specific implementation detail of MiLoRE in this work is proved to be the proper choice. In ablation settings D and E, the mixture of low-rank experts (MiLoRE) is respectively replaced with mixture of MLP experts (MoE) and DV-LoRA from \cite{endodac} to prove the outstanding contribution of MiLoRE proposed in this work. At last, the effect of intrinsic image alignment (IIA) is also demonstrated with ablation study F.

\begin{table}[!ht]
\caption{Ablation results on SCARED dataset and Hamlyn dataset. 'Ours' denotes the proposed MiLoRE. 'Ours-m-n' and 'MoE-m-n' denotes that the number of experts is \textbf{m}, while top-k is set to \textbf{n}. The {\color[HTML]{FF0000}best results} are in red.}
\scriptsize
\centering
{\fontsize{8}{10}\selectfont
\begin{tabular}{l|cc|c|ccccc}
\hline
                   & \multicolumn{2}{c|}{\textbf{PEFT}}            &                                & \multicolumn{5}{c}{\textbf{SCARED(Trained)   / Hamlyn(Zero-shot)}}                                                                                                                   \\ \cline{2-3} \cline{5-9} 
\multirow{-2}{*}{} & \textbf{QKV}          & \textbf{FFN}          & \multirow{-2}{*}{\textbf{IIA}} & \textbf{$Rel_{Abs}\downarrow$}     & \textbf{$Rel_{Sq}\downarrow$}      & \textbf{$RMSE\downarrow$}           & \textbf{$RMSE_{Log}\downarrow$}    & \textbf{$\delta\uparrow$}          \\ \hline
A                  & Ours-3-2            & Ours-5-2            & \CheckmarkBold                 & {\color[HTML]{FF0000} 0.047/0.140} & {\color[HTML]{FF0000} 0.315/3.087} & {\color[HTML]{FF0000} 4.246/11.700} & {\color[HTML]{FF0000} 0.067/0.174} & {\color[HTML]{FF0000} 0.984/0.813} \\
B                  & Ours-4-2            & Ours-5-2            & \CheckmarkBold                 & 0.051/0.153                        & 0.348/3.672                        & 4.428/12.721                        & 0.072/0.189                        & 0.980/0.792                        \\
C                  & Ours-4-2            & Ours-4-2            & \CheckmarkBold                 & 0.052/0.156                        & 0.392/3.973                        & 4.666/12.941                        & 0.074/0.192                        & 0.980/0.792                        \\
D                  & MoE-3-2               & MoE-5-2               & \CheckmarkBold                 & 0.081/0.164                        & 0.836/3.818                        & 6.892/13.526                        & 0.115/0.201                        & 0.929/0.760                        \\
E                  & DV-LoRA & DV-LoRA & \CheckmarkBold                 & 0.050/0.151 & 0.334/3.639 & 4.342/12.469 & 0.071/0.186 & 0.983/0.798 \\
F                  & Ours-3-2            & Ours-5-2            & \XSolidBrush                   & 0.052/0.149                      & 0.366/3.545                       & 4.578/12.378                        & 0.074/0.186                        & 0.979/0.797                        \\ \hline
\end{tabular}}
\label{ab}
\end{table}

\begin{table}[!ht]
\scriptsize
\centering
\caption{Results of ego-motion estimation and camera intrinsics prediction. \textbf{The best results} are in bold, while \underline{the second best results} are underlined.}
{\fontsize{8}{10}\selectfont
\begin{tabular}{c|ccccc|cccc}
\hline
\multirow{2}{*}{\textbf{Methods}} & \multicolumn{5}{c|}{\textbf{Absolute Trajectory Error}}                                                 & \multicolumn{4}{c}{\textbf{Camera Intrinsics $Rel_{Abs}$}}        \\ \cline{2-10} 
                                  & \textbf{Seq. 1} & \textbf{Seq. 2} & \textbf{Seq. 3} & \textbf{Seq. 4} & \textbf{Seq. 5} & \textbf{fx} & \textbf{fy} & \textbf{cx} & \textbf{cy} \\ \hline
AF-SfMLearner\cite{afsfm}         & 0.0304          & 0.0941    & 0.0841          & \textbf{0.0742} & \underline{0.0682}    & \multicolumn{4}{c}{Provided}                          \\
IID-SfMLearner\cite{iidsfm}         & \underline{0.0296}        & 0.0951    & 0.0851          & 0.0764 & \underline{0.0682}   & \multicolumn{4}{c}{Provided}                          \\
EndoDAC\cite{endodac}             & \textbf{0.0290} & \underline{0.0936}          &\underline{0.0832}    & 0.0776          & 0.0687          & 5.00\%      & 5.67\%      & 1.66\%      & 2.02\%      \\
EndoMINI(Ours)                    &0.0302    & \textbf{0.0927} & \textbf{0.0814} & \underline{0.0754}    & \textbf{0.0678} & \textbf{0.51\%}      & \textbf{0.31\%}      & \textbf{1.02\%}      & \textbf{0.98\%}     \\ \hline
\end{tabular}}
\label{ego}
\end{table}

\subsection{Ego-motion and Camera Intrinsics Estimation}
In the experiments, five different sequences from SCARED dataset are selected to compare results of ego-motion estimation from the proposed method and existing methods. The results of existing works are obtained based on the model with publicly provided parameters. As Table \ref{ego} shows, the proposed method can perform the most accurate ego-motion estimation without given camera intrinsics. Furthermore, for the prediction of camera intrinsics, our method is compared with EndoDAC \cite{endodac} on these five sequences. As Table \ref{ego} shows, with self-supervised learning based on the proposed framework, the network can provide more accurate camera intrinsics, which can also demonstrate the outstanding performance and generalization of our method.

\section{Conclusion}
In this work, a novel self-supervised depth estimation framework for endoscopic scenes, EndoMINI, is proposed to boost the generalization of the model. Based on features from various endoscopic images, expert networks can be selected for weighted inference in the proposed method based on mixture of low-rank experts. Moreover, intrinsic image decomposition which can filter the illumination influence is introduced into the proposed framework to deal with various illumination conditions. Experimental results demonstrate the state-of-the-art performance of the proposed method for both supervised depth estimation and zero-shot depth estimation with high generalizability in endoscopy. With high-quality 3D perception, this work can promote the development of minimally invasive surgery, especially for robot-assisted surgery.

\begin{credits}
\subsubsection{\ackname} This work was supported by Ministry of Science and Technology (MOST) of China Key Project  2025YFE0122500, Shenzhen-Hong Kong-Macau Technology Research Programme (Type C) STIC Grant SGCX20250526153900001, Hong Kong RGC CRF C4026-21G, RIF R4020-22, GRF 14211420, 14216022 \& 14203323.
\subsubsection{\discintname}
The authors have no competing interests to declare that are relevant to the content of this article.
\end{credits}

%
%
%
\bibliographystyle{splncs04}
\bibliography{refs}

\end{document}